# Specification-delta-driven data governance: an empirical study of the «spec-delta» as the unit of change in lakehouse data platforms

**Pablo Ramírez Amador, PhD Candidate**

PhD Candidate in Computer Science — Universidad Abierta Interamericana (UAI)

Zaragoza, Spain · Working version (preprint) · August 2026

## Article info

## Abstract

Spec-Driven Development (SDD) has consolidated the idea that the specification —rather than the code— should be the primary artefact governing AI-assisted work. Tools such as GitHub Spec Kit, and proposals such as Constitutional SDD, have formalised this principle in the software domain, while the executable data-contracts literature has extended it to schema and quality enforcement at run time. Nevertheless, the treatment of the specification delta —OpenSpec's core idea that every change should produce a reviewable increment of requirements— as the unit of change in data platforms remains empirically unexplored, even though many data-platform changes are contractual (new datasets, service-level agreements, metric semantics, access policies) rather than purely code changes. This work formalises the spec-delta concept, proposes a taxonomy of data-platform changes according to their suitability for incremental specification, and defines a controlled experiment comparing a spec-delta-driven workflow against a conventional code pull-request workflow without a delta. The response variables are discovery-to-deployment time, the density of defects reaching the Silver and Gold lakehouse layers, cross-tool metric divergence, and reviewer cognitive load measured with NASA-TLX. The paper explicitly reserves a demonstration-and-laboratory section for instantiation on a real lakehouse environment. The contribution is not a tool but reproducible evidence and an applicability guide that helps to avoid the up-front over-specification antipattern.

## Corresponding author

**Pablo Ramírez Amador**

PhD in Computer Science — Universidad Abierta Interamericana (UAI), Argentina.

Zaragoza, Spain.

## 1. Introduction

The mass adoption of code-generation assistants and autonomous agents has shifted the software-engineering bottleneck away from writing code and towards specifying and governing it. The movement known as Spec-Driven Development (SDD) responds to this tension by proposing that the primary artefact should no longer be the code but a versioned, reviewable and executable specification, from which both the implementation and its acceptance tests are derived. GitHub Spec Kit realised this workflow as a pipeline of phases —constitution, specification, plan, tasks and implementation— in which each phase produces a Markdown artefact consumed by the next and subject to human review (GitHub, 2025).

In parallel, the data-platform ecosystem has converged on the medallion architecture —the Bronze, Silver and Gold layers that denote increasing levels of quality— as the dominant pattern for organising the lakehouse (Databricks / Microsoft Learn, 2026). A large proportion of the changes introduced into a data platform are not code changes in the strict sense but contractual changes: the addition of a new dataset, the modification of a service-level agreement (SLA), the redefinition of the semantics of an executive metric, or the adjustment of an access policy. Such changes have downstream consumers, traceability obligations and a blast radius that reaches beyond the file that was modified.

Recent literature has examined two extremes of the spectrum in detail. At one extreme, SDD applied to code, including its Constitutional SDD variant, which embeds non-negotiable security constraints in the specification layer with traceability down to the line of code (Marri, 2026). At the other extreme, executable data contracts, which formalise agreements on schema, semantics and quality between producers and consumers and make them enforceable at run time (Bhoite, 2025). Between the two lies a gap: no one has studied empirically what happens when OpenSpec's core idea —that every change should produce a readable requirements delta— is adopted as the atomic unit of change and review in a data platform.

### 1.1. Problem and motivation

The problem we address is twofold. First, a problem of change legibility: in conventional workflows the reviewer must reconstruct, from a code pull request and the transformation logic, which platform capability is changing, which consumers are affected and which guarantees are preserved or broken. Second, a problem of verifiable governance: without a requirements artefact attached to the change, it is difficult to make the promotion of data to Silver or Gold conditional on explicit criteria of quality, ownership and traceability. The motivating hypothesis is that a well-formed spec-delta —with SHALL requirements and GIVEN/WHEN/THEN scenarios— reduces the reviewer's comprehension effort and the leakage of defects into higher layers, without necessarily incurring the cost of the up-front over-specification antipattern that practice has flagged as a risk of SDD.

### 1.2. Contributions

This article makes four contributions:

- **A formalisation of the spec-delta concept** as the unit of change in data platforms, with its anatomy (SHALL requirements, GIVEN/WHEN/THEN scenarios, architecture decisions and a validation checklist) and its life cycle tied to the lakehouse promotion gates.
- **A taxonomy of data-platform changes** that classifies which types of change benefit from incremental specification and which do not, in order to bound applicability and avoid over-specification.
- **A controlled, reproducible experimental design** that compares a spec-delta-driven workflow against a conventional workflow, with operational response variables: discovery-to-deployment time, defect density in Silver/Gold, cross-tool metric divergence between BI tools, and reviewer cognitive load (NASA-TLX).
- **A reserved demonstration-and-laboratory section** with methodological scaffolding to instantiate the experiment on a real lakehouse environment, together with data-collection and results-reporting templates.

### 1.3. Structure of the article

Section 2 reviews the background and related work and delimits the research gap. Section 3 presents the conceptual framework of the spec-delta and its associated taxonomy. Section 4 states the research questions and hypotheses. Section 5 details the experimental design. Section 6 reserves and describes the demonstration and laboratory. Section 7 analyses the threats to validity. Section 8 discusses the implications and Section 9 concludes and outlines future work.

## 2. Background and related work

### 2.1. Spec-Driven Development and OpenSpec

SDD inverts the traditional relationship between specification and code: the specification ceases to document the code and becomes the source that generates and constrains it. GitHub Spec Kit structures this process around a constitution of non-negotiable principles and a sequence of reviewable artefacts (specification, plan and tasks) before implementation (GitHub, 2025). OpenSpec's contribution to this picture is conceptually subtle but significant (OpenSpec, 2026): instead of maintaining a monolithic specification, it proposes that each change should produce a specification delta with its own requirements and scenarios, so that the reviewer understands which capability is changing without reconstructing it from the code. The Constitutional SDD variant —which adapts the notion of Constitutional AI (Bai et al., 2022)— takes the approach into the security domain, encoding constraints derived from CWE/MITRE with MUST/SHOULD/MAY requirement levels (Bradner, 1997) and traceability down to the implementation, and reporting substantial reductions in security defects in a case study of banking microservices (Marri, 2026).

Professional practice has warned, nonetheless, of a risk inherent in SDD: a bias towards excessive up-front specification and towards large «big-bang» deliveries, catalogued as an antipattern in industry technology radars (ThoughtWorks, 2025). This work takes that warning seriously and turns it into an empirical question: not whether SDD is good or bad in the abstract, but for which classes of data change the spec-delta delivers net value.

### 2.2. Executable data contracts and quality in the lakehouse

Data contracts formalise agreements between producers and consumers on schema, semantics and quality expectations. Recent research has proposed generating them with the assistance of language models —for example, through fine-tuning with LoRA/PEFT to produce valid definitions in JSON Schema or Avro integrated into platforms such as Databricks and Snowflake— and has measured notable reductions in manual effort while also flagging challenges of hallucination, versioning and maintenance (Bhoite, 2025). The concern for data quality as a source of silent failure is well founded: work on data smells catalogues suspicious patterns in the data of AI-based systems (Foidl et al., 2022), and studies on underspecification show how the lack of specification compromises the credibility of machine-learning models (D'Amour et al., 2020).

The essential difference from our approach is the locus of control. Data contracts operate predominantly at run time and at the schema/quality level of an already-defined data product; the spec-delta operates at review time, as an increment of requirements that governs the decision to admit or promote a change. The two are complementary: the spec-delta may require, among its criteria, the existence and verification of a data contract.

### 2.3. The medallion architecture as a per-layer contract

The medallion architecture, characteristic of lakehouse platforms (Armbrust et al., 2020, 2021; Delta Lake, 2026), describes a series of data layers denoting increasing quality: Bronze preserves the raw ingestion with minimal validation, Silver applies cleansing, validation and deduplication, and Gold publishes aggregations and dimensional models ready for BI, machine learning and operations (Databricks /

Microsoft Learn, 2026). Read as a specifiable pattern, each boundary between layers is a natural gating point: what is preserved in Bronze, what is validated in Silver and what is published in Gold can be expressed as verifiable requirements. This reading turns the promotion between layers into the ideal place to require an approved spec-delta as a precondition.

### 2.4. Semantic layer and metric divergence

The proliferation of duplicated metric definitions across different BI tools produces divergences in figures that ought to be identical. Semantic layers centralise the definition of metrics over existing models and delegate joins, permissions and integrations downstream (dbt Labs, 2026), turning metrics such as revenue, churn or conversion into versionable objects. In our framework, the semantics of an executive metric is precisely one of the contractual changes that a spec-delta should capture, with its definition, permitted dimensions, granularity, owner and example queries.

### 2.5. Traceability and blast radius

Lineage traceability —captured through open specifications of run events with dataset, job and run entities (OpenLineage, 2026)— makes it possible to compute the blast radius of a change before it reaches production and to reconstruct the chain of custody after an incident. In the experimental design, the emitted lineage is both a criterion on the spec-delta's validation checklist and an instrument for measuring the real impact of each change.

### 2.6. Delimitation of the research gap

In summary: there is evidence on SDD for code and on data contracts at run time, but there is no empirical evidence on the spec-delta as the unit of change and review in data platforms, nor a taxonomy that disciplines its applicability so as to avoid over-specification. This article sits exactly in that gap, with a contribution that is empirical rather than instrumental in nature.

## 3. Conceptual framework: the spec-delta as the unit of change

### 3.1. Definition

We define a spec-delta as the minimal, self-contained increment of specification associated with a change to a data platform, written in such a way that a reviewer can decide on its admission without reconstructing the intent from the code. A spec-delta describes the difference in capability —what is added, modified or removed with respect to the current specified state— and not the whole system. Its unit of analysis is the change, not the complete data product —the unit of the data-mesh paradigm (Dehghani, 2022)— which distinguishes it both from a monolithic requirements document and from a static data contract.

### 3.2. Anatomy of a spec-delta

A well-formed spec-delta is composed of five elements:

- **SHALL requirements.** Normative statements, with requirement levels according to RFC 2119 (MUST/SHOULD/MAY; Bradner, 1997), that describe the guarantees the change introduces or preserves over the data product.
- **GIVEN/WHEN/THEN scenarios.** Executable acceptance criteria that fix observable behaviour —for example, quality conditions, deduplication rules or the semantic invariants of a metric— and that serve as the basis for the tests.
- **Architecture decisions (ADR).** A concise record of the decision taken, its alternatives and its justification, so as to preserve the traceability of the reasoning.
- **Validation checklist.** A set of objective checks —quality tests passed, lineage emitted, owner defined, downstream impact computed, reproducible execution evidence— that condition promotion.

- **Expected artefacts.** Examples of the products that accompany the change: the YAML definition of the dataset, the metric definition in the semantic layer, the quality-expectations suite and the lineage event.

Applying Meyer's principle of Design by Contract (Meyer, 1992) to the data domain, the SHALL requirements and the GIVEN/WHEN/THEN scenarios act as verifiable preconditions and postconditions of the data product's state transition between layers.

### 3.3. Life cycle and promotion gates

The spec-delta is anchored to the boundaries of the medallion architecture. The governance rule we propose is explicit: no change is promoted to Silver or Gold without an approved spec-delta, quality tests passed, lineage emitted, owner defined, downstream impact computed and reproducible execution evidence. The gate does not replace human review; it makes it more efficient by presenting the change as a difference in requirements rather than as a code diff.

*Figure 1 (placeholder). Life cycle of the spec-delta across the Bronze→Silver→Gold promotion gates. [Insert diagram in the laboratory instantiation.]*

### 3.4. Taxonomy of data-platform changes

To discipline applicability and avoid over-specification, we classify changes according to their nature and their expected suitability for the spec-delta. Table 1 summarises the proposed taxonomy, which itself constitutes a hypothesis to be tested in the laboratory.

| Change class | Examples | Nature | Suitability |
|---|---|---|---|
| New data product | Registering a dataset Bronze→Silver→Gold with consumers | Contractual | High |
| Metric semantics | Redefining revenue, churn or conversion in the semantic layer | Contractual | High |
| Data SLA / SLO | Changing committed freshness, availability or completeness | Contractual | High |
| Access policy | Masking, column-level RBAC, retention | Contractual | High |
| Schema evolution | Adding/renaming columns, type changes | Mixed | Medium |
| Quality rule | New expectation, deduplication threshold | Mixed | Medium |
| Internal refactor | Rewriting a transformation without a contract change | Code | Low |
| Performance optimisation | Repartitioning, Z-order, cluster tuning | Code | Low |

*Table 1. Taxonomy of data-platform changes and expected suitability for incremental specification (spec-delta). The «Suitability» column expresses a hypothesis to be tested, not a result.*

## 4. Research questions and hypotheses

The study is organised around four research questions (RQ) and their associated hypotheses, formulated following the Goal-Question-Metric paradigm (Basili et al., 1994).

- **RQ1.** Does the spec-delta-driven workflow reduce discovery-to-deployment time compared with a conventional workflow based on a code pull request without a delta?
- **RQ2.** Does it reduce the density of defects that reach the Silver and Gold layers?
- **RQ3.** Does it reduce the divergence of one and the same metric computed in different BI tools?
- **RQ4.** Does it change the reviewer's cognitive load, and in which direction?

The corresponding null (H0) and alternative (H1) hypotheses are summarised in Table 2.

| RQ | Null hypothesis (H0) | Alternative hypothesis (H1) |
|---|---|---|
| RQ1 | There is no difference in discovery-to-deployment time between conditions. | The spec-delta workflow reduces discovery-to-deployment time. |
| RQ2 | There is no difference in defect density in Silver/Gold. | The spec-delta workflow reduces defect density in Silver/Gold. |
| RQ3 | There is no difference in metric divergence between BI tools. | The spec-delta workflow reduces metric divergence. |
| RQ4 | There is no difference in the reviewer's cognitive load (NASA-TLX). | The spec-delta workflow changes the reviewer's cognitive load. |

*Table 2. Research questions and hypotheses. RQ4 is formulated as two-tailed because it does not presuppose the direction of the effect on cognitive load.*

## 5. Experimental design

The study is conceived as a controlled experiment in software engineering, following the consolidated methodological guidelines for experimentation (Wohlin et al., 2012) and for case studies (Runeson & Höst, 2009). A design that controls for variability between subjects and between tasks is preferred.

### 5.1. Design and conditions

Two experimental conditions are considered: (C1) the conventional workflow, in which each change is materialised as a code pull request with its free-text description; and (C2) the spec-delta workflow, in which each change is accompanied by the specification increment described in Section 3. To mitigate the effect of individual differences, a within-subject, counterbalanced crossover design is proposed: each participant carries out equivalent tasks under both conditions, alternating the order between participants so as to neutralise learning and fatigue effects. As an alternative, where the organisational context permits, a between-subjects A/B design by teams with block randomisation is admitted.

### 5.2. Variables and metrics

The independent variable is the condition (C1 vs. C2). The dependent variables and their operationalisation are set out in Table 3.

| Dependent variable | Operationalisation | Instrument / source |
|---|---|---|
| Discovery-to-deployment | Time elapsed from the formulation of the need to the effective promotion to Gold. | Timestamps from the version-control system and the orchestrator. |
| Defect density | Defects detected in Silver/Gold per unit of change (or per 1,000 affected rows). | Quality suite, incidents and post-promotion verification. |
| Metric divergence | Relative difference of one and the same metric computed in ≥2 BI tools. | Queries compared against the semantic layer. |
| Cognitive load | Global score and sub-scale scores after each review task. | NASA-TLX questionnaire. |
| Blast radius | Number of downstream assets affected by the change. | Emitted lineage graph. |

*Table 3. Dependent variables, operationalisation and measurement instruments.*

### 5.3. Instrumentation

Discovery-to-deployment time is instrumented with automatic timestamps from the version-control system and the data orchestrator, avoiding self-reporting. Defect density is obtained from the quality suite and from an independent verification carried out after promotion. Metric divergence is computed by running the same semantic query in at least two BI tools. The reviewer's cognitive load is measured with the NASA-TLX index (Hart & Staveland, 1988), widely validated, administered immediately after each review task.

## 5.4. Population, sampling and experimental unit

The experimental unit is the change task. A bank of equivalent tasks derived from the taxonomy in Table 1 is built, balanced by class and by difficulty. The participants are data engineers and reviewers with comparable experience; the sample size will be determined by an a priori power analysis, fixing $\alpha = 0.05$ and a target power of 0.80 for a medium effect size, adjusting for the within-subject structure.

## 5.5. Procedure

- A homogeneous training session on both workflows and on the laboratory environment.
- Counterbalanced assignment of tasks and conditions.
- Execution of the tasks with automatic recording of timestamps and artefacts.
- Administration of NASA-TLX after each review task.
- Independent post-promotion verification and computation of metric divergence.
- A brief closing interview for complementary qualitative data.

## 5.6. Statistical analysis

For continuous variables with a within-subject design, linear mixed models with the subject as a random effect will be used, or paired tests (paired Student's t or Wilcoxon depending on normality, verified with Shapiro-Wilk). For defect counts, Poisson or negative-binomial regression will be considered. Effect sizes (Cohen's d or r) and confidence intervals will be reported, and a correction for multiple comparisons will be applied. All the material —task bank, instrumentation scripts and anonymised data— will be published to support reproducibility.

# 6. Demonstration and laboratory

## 6.1. Reference environment

The reference instantiation runs on an Azure Databricks workspace with Unity Catalog enabled, taking the Global Superstore dataset organised in a medallion architecture as its domain (catalogue adbs_poc_vass; bronze, silver and gold schemas). Table 5 lists the components and their role; the exact build numbers are pinned in the reproducibility notebook (6.5) to guarantee re-execution.

| Component | Technology | Role in the laboratory |
|---|---|---|
| Lakehouse engine and table format | Azure Databricks + Delta Lake on Unity Catalog | ACID storage and catalogue governance (adbs_poc_vass; bronze/silver/gold schemas) |
| Data domain | Global Superstore (medallion Bronze→Silver→Gold) | Bank of data products on which the changes are applied |
| Orchestrator (asset-based) | Databricks Asset Bundles + Databricks Jobs | Declarative deployment and execution; promotion timestamps |
| Quality framework | Great Expectations (per-layer expectations) | Suite that conditions promotion to Silver/Gold |
| Semantic layer | dbt Core + dbt Semantic Layer (MetricFlow) | Single, versionable definition of metrics (revenue, profit_margin) |
| Lineage emitter | OpenLineage (Spark/Databricks) and Unity Catalog lineage | Blast radius and change traceability |
| Version control | Git (GitHub) | Discovery/review timestamps; PR (C1) and spec-delta (C2) |
| BI tools (≥2) | Power BI and Microsoft Fabric | Computation of the divergence of one and the same metric |
| Gate harness | OpenSpec framework (hexagonal architecture) | Generation and validation of spec-deltas; promotion gates |

*Table 5. Components of the reference laboratory environment and their role. The exact versions are pinned in the reproducibility repository (§6.5).*

*Figure 2. Architecture of the laboratory environment: Git → Databricks Asset Bundles → medallion (Delta Lake/Unity Catalog) with Great Expectations expectations at the Silver/Gold promotion gates, dbt semantic layer (profit_margin, revenue), OpenLineage lineage emission and consumption from Power BI and Microsoft Fabric for the divergence computation. [Insert architecture diagram in the laboratory instantiation.]*

## 6.2. Change-task bank

Document the set of equivalent tasks used, derived from the taxonomy in Table 1, indicating class, estimated difficulty and equivalence criteria between conditions.

The bank comprises eight base tasks (T1–T8), one for each class in Table 1, defined over Global Superstore (Table 6). For the within-subject design, each task Ti has an equivalent twin Ti′ of the same class and difficulty (for example, a product registration for another region): each participant carries out one under condition C1 and its twin under C2, with counterbalanced order, so that no specific task is repeated for the same subject. The «Cons.» column indicates the expected downstream consumers, which fix the expected blast radius.

| ID | Class (Table 1) | Task on Global Superstore | Diff. | Cons. |
|---|---|---|---|---|
| T1 | New data product | Registering the product gold.sales_by_region (Bronze→Silver→Gold) | High | 3 |
| T2 | Metric semantics | Redefine profit_margin in the semantic layer (margin on sales net of returns) | High | 4 |
| T3 | Data SLA / SLO | Reduce the committed freshness of silver.orders from 24 h to 6 h | Medium | 3 |

| ID | Class (Table 1) | Task on Global Superstore | Diff. | Cons. |
|---|---|---|---|---|
| T4 | Access policy | Dynamic data masking of customer data and column-level RBAC in Unity Catalog | Medium | 2 |
| T5 | Schema evolution | Add the discount_band column to silver.order_items | Medium | 2 |
| T6 | Quality rule | New uniqueness/deduplication expectation by order_id in silver.orders | Low | 1 |
| T7 | Internal refactor | Rewrite the Bronze→Silver transformation of returns without a contract change | Low | 0 |
| T8 | Performance optimisation | OPTIMIZE/Z-order and repartitioning of gold.sales_agg | Low | 0 |

*Table 6. Change-task bank derived from the taxonomy (Table 1), instantiated on Global Superstore. Each task has an equivalent twin for the within-subject counterbalancing.*

## 6.3. Laboratory-session protocol

Each participant is randomly assigned to one of two counterbalancing orders (order A: C1→C2; order B: C2→C1), following a homogeneous training session on both workflows and on the environment. The number of participants is fixed by the a priori power analysis of §5.4 ($\alpha = 0.05$; power 0.80; medium effect; adjustment for the within-subject structure).

Per-participant isolation is achieved with a dedicated schema in Unity Catalog (for example adbs_poc_vass.exp_<id>) and a participant's own Git branch, so that tasks do not interfere with one another. Timestamps are captured automatically: the start (discovery) from the first commit of the task and the end from the execution mark of the promotion job in Databricks, avoiding self-reporting. Metric divergence is obtained by running the same semantic query in Power BI and in Microsoft Fabric; NASA-TLX is administered digitally immediately after each review task.

The criterion for «effective promotion» —which marks the end of the discovery-to-deployment time— is the moment at which the target Gold table is published in the catalogue satisfying the gate of §3.3: Great Expectations suite green, OpenLineage event emitted and owner assigned. In C1 the promotion is anchored to the merge of the pull request and the correct execution of the job; in C2, to the approval of the spec-delta and that same execution. An independent post-promotion verification counts the defects that reach Silver/Gold.

## 6.4. Results templates

The templates for the results tables that are to be populated with the laboratory data are offered below.

| Variable | C1 (conventional) | C2 (spec-delta) | p / d |
|---|---|---|---|
| Discovery-to-deployment (h) | — | — | — |
| Defects in Silver/Gold | — | — | — |
| Metric divergence (%) | — | — | — |
| NASA-TLX (global) | — | — | — |

*Table 4 (template). Aggregated results by condition. To be populated with means/medians, dispersion, p-value and effect size.*

*Figure 3 (template). Box plots by condition (C1 vs. C2) for each variable in Table 4, plus a panel disaggregated by change class (Table 1) contrasting the suitability hypothesis: high expected benefit for contractual changes (T1–T4), decreasing for the mixed ones (T5–T6) and null or negative for the code ones (T7–T8). Structure without real data: to be populated after the laboratory has been run.*

## 6.5. Reproducibility notebook

The material is organised in a Git repository with the following structure: openspec/ (constitution, spec-delta templates and gate definitions), tasks/ (the T1–T8 bank and its twins, with equivalence criteria), instrumentation/ (scripts for capturing timestamps from Git and Databricks Jobs, and for computing Power BI/Fabric divergence), expectations/ (per-layer Great Expectations suites), lineage/ (OpenLineage configuration) and data/ (an anonymised extract of Global Superstore and aggregated results). The public link and the DOI will be added upon publication of the laboratory, in line with the «Data and materials availability» section.

### 6.6. Illustrative use case: the add-profit-margin-contract spec-delta

As a concrete demonstration, Table 7 shows a complete spec-delta for task T2 (metric semantics), written with the template from Appendix A. The change redefines the profit_margin metric over Global Superstore as the quotient of profit and sales net of returns, and is governed at the promotion gate to Gold.

| Identifier | add-profit-margin-contract |
|---|---|
| Motivation | Unify the profit margin as a single, versionable metric, eliminating divergent definitions between Power BI and Fabric over Global Superstore. |
| SHALL requirements | The system SHALL expose profit_margin in the semantic layer as profit / net_sales; SHALL exclude returned lines (returns) from the denominator; SHOULD document owner and granularity (by order, region and category); MAY publish example queries. |
| Scenarios | GIVEN an order with a partial return, WHEN profit_margin is computed, THEN net_sales discounts the returned amount and the margin reflects the net profit. GIVEN two BI tools, WHEN they query the same metric and period, THEN the target relative divergence is 0 %. |
| ADR | The metric is centralised in the dbt Semantic Layer (rejected alternative: defining it in each BI report). Justification: single source of truth and traceability; cost: a dependency on the deployment of the semantic layer. |
| Checklist | Quality tests ☑ · Lineage emitted ☑ · Owner ☑ (Sales domain) · Downstream impact ☑ (4 consumers) · Reproducible evidence ☑ |
| Artefacts | metrics/profit_margin.yml (dbt) · net_sales Great Expectations suite · OpenLineage event of the gold.sales_agg job · Power BI/Fabric verification query |

*Table 7. Complete spec-delta of the illustrative use case (task T2), written with the template from Appendix A.*

## 7. Threats to validity

### 7.1. Construct validity

The operationalisation of «change legibility» through cognitive load and defect density is indirect. It is mitigated by triangulating NASA-TLX with objective process data and with qualitative evidence from the closing interview.

### 7.2. Internal validity

Learning and fatigue effects are the main threat of the within-subject design; they are controlled through counterbalancing of the order of conditions and through rest breaks. Automatic instrumentation reduces the self-reporting bias of the time measurements.

### 7.3. External validity

The results obtained in a laboratory environment may not generalise to all organisations or to all lakehouse engines. The taxonomy in Table 1 bounds the scope and allows reporting by change class, favouring nuanced transferability.

### 7.4. Conclusion validity

The sample size and the choice of tests condition the reliability of the inferences; this is mitigated with an a priori power analysis, the reporting of effect sizes and confidence intervals, and a correction for multiple comparisons.

## 8. Discussion

The expected value of the spec-delta does not lie in specifying more, but in specifying the change with just enough granularity for review and promotion to be informed decisions. The proposed taxonomy anticipates that the benefit will be concentrated in contractual changes —registrations of data products, metric semantics, SLA/SLO and access policies— and will be attenuated, or even reversed, in internal refactorings and performance optimisations, where the cost of writing the delta might not pay off. Should this be confirmed, the pattern would offer a practical rule for avoiding the over-specification antipattern: reserve the spec-delta for changes that alter contracts observable by consumers.

The approach is complementary, not competitive, with executable data contracts and with the AI-assisted generation of contracts (Bhoite, 2025): the spec-delta may require, among its validation criteria, the existence and verification of a data contract, and may itself be generated with the assistance of language models, provided that its approval remains under human review. Likewise, the proposed promotion gate fits naturally with architectures of data agents connected through open protocols (Model Context Protocol, 2026) that can only propose specs and changes, never deploy directly.

## 9. Conclusions and future work

This article has formalised the spec-delta as the unit of change in data platforms, has proposed a taxonomy of applicability and has defined a controlled, reproducible experimental design to test its effect on discovery-to-deployment time, the leakage of defects into Silver/Gold, metric divergence and reviewer cognitive load. The central contribution is empirical in nature: not a tool, but a framework of evidence that makes it possible to decide when incremental specification delivers net value.

Future work includes the execution of the laboratory described in Section 6, the cross-validation of the taxonomy across more than one organisation and more than one lakehouse engine, the integration of the spec-delta with automatic blast-radius detection over the lineage graph, and the study of language-model assistance in the writing of deltas under human-review gates.

### Data and materials availability

The reproducibility materials (task bank, instrumentation scripts, spec-delta templates and anonymised data) will be published in an open repository once the laboratory in Section 6 has been run.

## Appendix A. Spec-delta template

A minimal template for writing a spec-delta in the experimental workflow is offered. The fields in square brackets are completed per change.

| **Identifier** | **add-<short-capability> (for example, add-sales-metric-contract)** |
|---|---|
| Motivation | [Analytical or operational need that gives rise to the change] |
| SHALL requirements | SHALL/MUST/SHOULD/MAY … [guarantees introduced or preserved] |
| Scenarios | GIVEN … WHEN … THEN … [verifiable acceptance criteria] |
| ADR | [Decision taken, alternatives and justification] |
| Checklist | Quality tests ☐ · Lineage emitted ☐ · Owner ☐ · Downstream impact ☐ · Reproducible evidence ☐ |
| Artefacts | [Dataset YAML · semantic-layer metric · quality suite · lineage event] |

*Table A.1. Minimal spec-delta template for the experimental workflow.*